\documentclass{article}

\PassOptionsToPackage{numbers, square, compress}{natbib}
\usepackage[final]{tackling_climate_workshop_style}

\usepackage[utf8]{inputenc}
\usepackage[T1]{fontenc}
\usepackage[hidelinks]{hyperref}
\usepackage{url}
\usepackage{booktabs}
\usepackage{amsfonts}
\usepackage{graphicx}
\usepackage{nicefrac}
\usepackage{microtype}
\usepackage[utf8]{inputenc}
\DeclareUnicodeCharacter{2212}{-}
\usepackage{graphicx}

\title{Weather Maps as Tokens: Transformers for Renewable Energy Forecasting}

\author{%
 Federico Battini\thanks{Corresponding author}\\
 IFAB Foundation\\
 Bologna, Italy\\
 \texttt{federico.battini@ifabfoundation.org}
 }

\begin{document}

\maketitle

\begin{abstract}
Accurate renewable energy forecasting is essential to reduce dependence on fossil fuels and enabling grid decarbonization. However, current approaches fail to effectively integrate the rich spatial context of weather patterns with their temporal evolution. This work introduces a novel approach that treats weather maps as tokens in transformer sequences to predict renewable energy. Hourly weather maps are encoded as spatial tokens using a lightweight convolutional neural network, and then processed by a transformer to capture temporal dynamics across a 45-hour forecast horizon. Despite disadvantages in input initialization, evaluation against ENTSO-E operational forecasts shows a reduction in RMSE of about 60\% and 20\% for wind and solar respectively. A live dashboard showing daily forecasts is available at:
\url{https://www.sardiniaforecast.ifabfoundation.it}.
\end{abstract}

\section{Introduction}

The transition to renewable energy sources is one of the most critical challenges in addressing climate change. Recent analyses demonstrate that the deployment of renewable energy is now driven by economics, with electricity potentially accounting for approximately 66\% of final energy consumption by mid-century in 1.5°C scenarios \cite{mercure2021,luderer2022}. However, this massive integration creates new grid management challenges because distributed renewable energy can substantially decrease grid reliability \cite{smith2022}.

Machine learning approaches have become essential tools for forecasting renewable energy, enabling more accurate predictions and facilitating grid integration. Significant advances in wind power forecasting have been achieved through attention-based architectures and adaptive calibration mechanisms \cite{belletreche2024,wang2025,keisler2024}. Meanwhile, solar power prediction has evolved toward sophisticated spatiotemporal architectures, with attention networks and graph neural networks capturing dynamic correlations between multiple sites \cite{liang2022,yang2025,jeon2022}. Hybrid CNN-LSTM architectures dominate high-performance applications, combining spatial feature extraction with temporal dependency modeling \cite{ladjal2025,bashir2025,ayene2024}. Transformer architectures represent an emerging frontier, with pure transformer models outperforming other approaches \cite{huang2023,oliveira2023,sakib2024}. The integration of numerical weather prediction (NWP) data and ensemble methods has further improved forecast accuracy through sophisticated preprocessing and multi-model combinations \cite{neumann2023,lafuente2025,wu2025}.

Despite these advances, existing CNN-Transformer models separate spatial and temporal features, limiting integration. This work introduces a novel paradigm that treats weather maps as tokens in transformer sequences, combining computer vision and natural language processing approaches for spatio-temporal prediction. Unlike existing methods, which handle spatial and temporal information separately, this approach unifies the processing of the complete meteorological context and leverages transformer attention mechanisms to model long-range dependencies.

\section{Methodology}

\subsection{Data}

This work uses the high-resolution NWP model MOLOCH \cite{tettamanti2002}, which covers Italy and provides two-day forecasts beginning at 3 a.m. Sardinia was chosen as the case study region. Meteorological inputs consist of hourly weather maps with a spatial resolution of 1.25 km (128×256 pixels after cropping). Wind forecasting employed horizontal wind components (U and V at 10 m), while solar forecasting used solar radiation, cloud cover, and temperature fields. Hourly renewable energy production and annual installed capacity data were obtained from the transparency platform \cite{entsoe2025} operated by the European Network of Transmission System Operators for Electricity (ENTSO-E).

The dataset spans from 2017 to 2024 and is divided into three periods: training (2017–2022), validation (2023), and testing (2024). Each sample consists of a 45-hour sequence of weather maps, enabling the identification of temporal dependencies beyond individual daily cycles. Data preprocessing included normalization computed exclusively from the training years to prevent leakage. Installed capacity was interpolated to hourly values for normalization.

\subsection{Model Architecture}

\begin{figure}[b]
\centering
\includegraphics[width=0.9\textwidth]{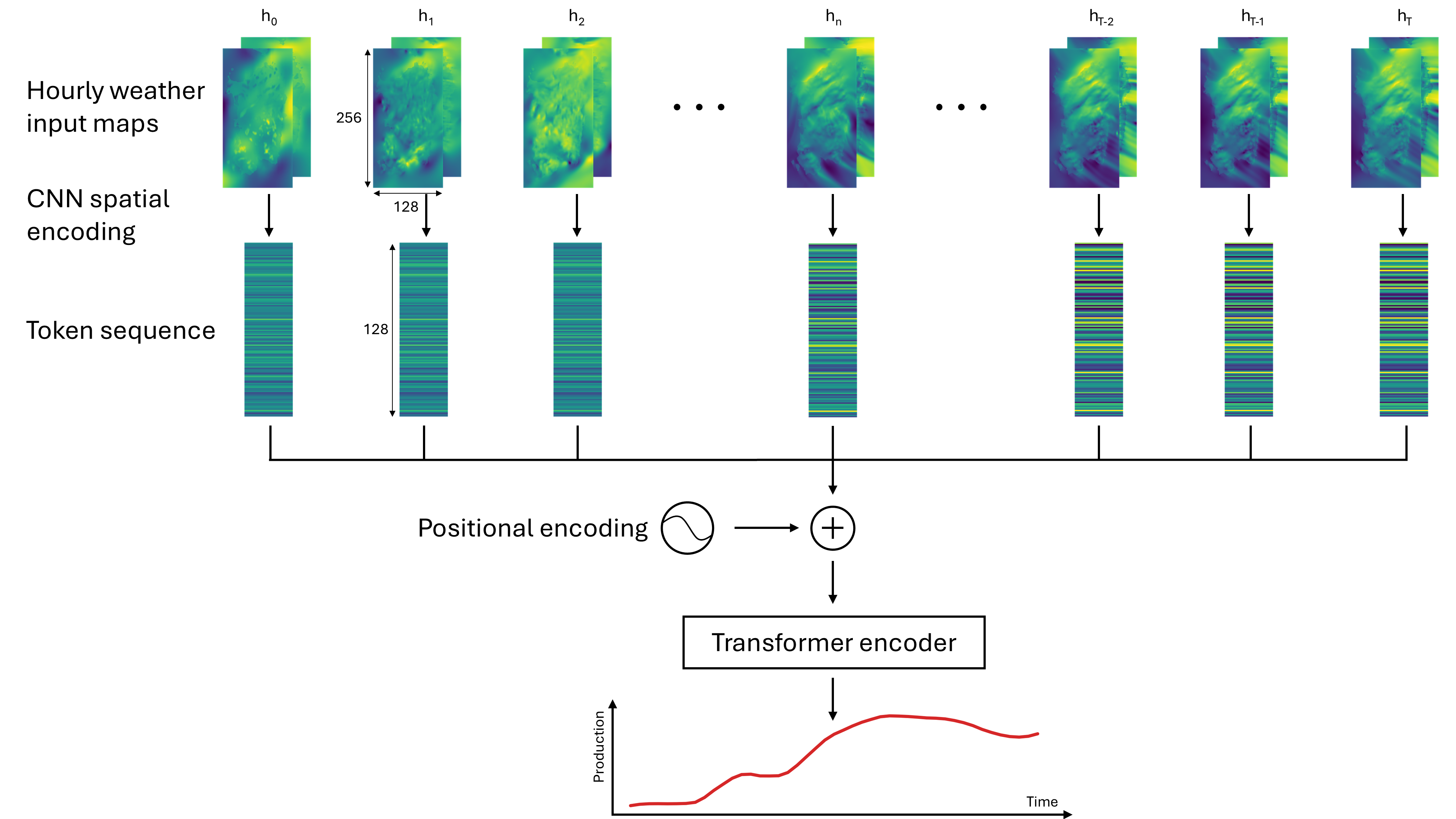}
\caption{Weather Maps as Tokens architecture.}
\label{fig:architecture}
\end{figure}

Figure \ref{fig:architecture} illustrates the proposed architecture. In the spatial encoding stage, each hourly weather input is processed independently through a lightweight 2D CNN encoder consisting of two convolutional layers with 16 and 32 filters, respectively, batch normalization and ReLU activation. Adaptive global average pooling handles the variable spatial dimensions while a fully connected layer maps the spatial features to 128-dimensional embeddings. The temporal modeling stage treats the sequence of 128-dimensional weather tokens, augmented with sinusoidal positional encodings, as input to a transformer encoder with two blocks, each using four-head attention and a feedforward network of hidden dimension 256. A linear prediction head then maps each temporal embedding to renewable energy production values, enabling hour-by-hour forecasting across the 45-hour horizon.

Training utilized PyTorch Lightning on a single NVIDIA A100 GPU with Adam optimizer (lr=$1\times{10^{-4}}$), MSE loss, and early stopping with patience equal to 20 epochs.

\section{Results}

\subsection{Model Performance}

\begin{figure}[b]
\centering
\begin{minipage}{0.48\textwidth}
\centering
\includegraphics[width=\textwidth]{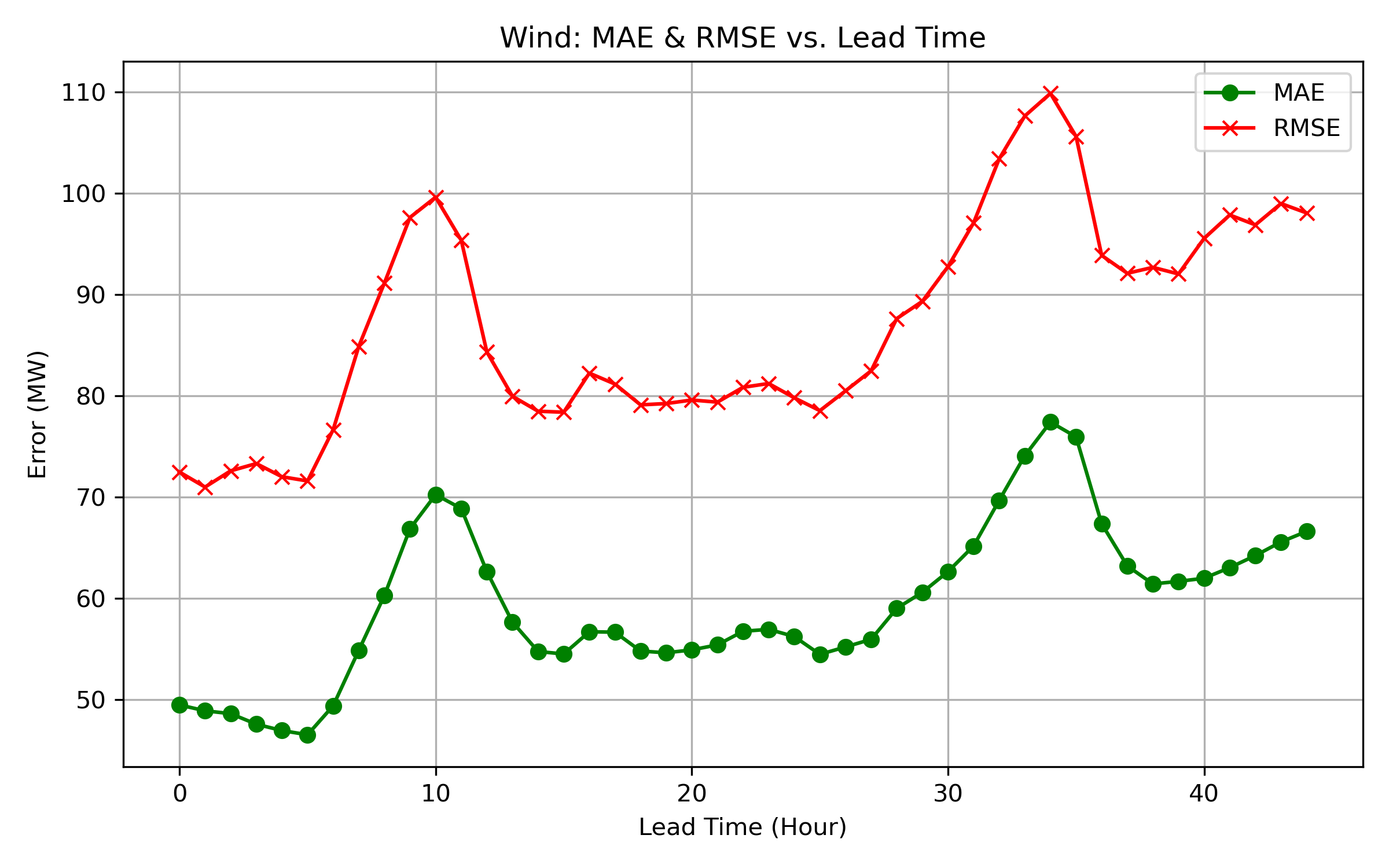}
\end{minipage}
\hfill
\begin{minipage}{0.48\textwidth}
\centering
\includegraphics[width=\textwidth]{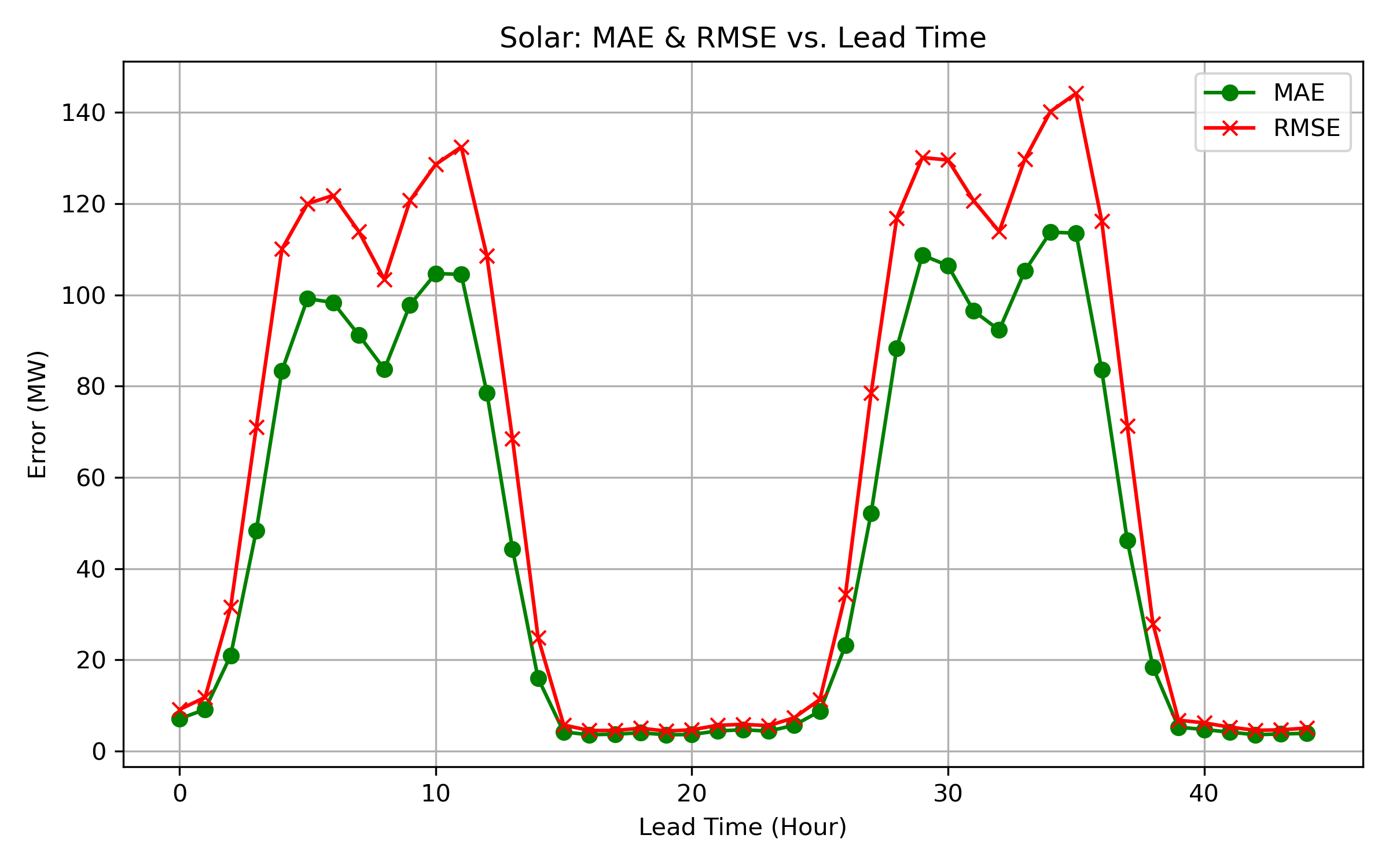}
\end{minipage}
\caption{MAE and RMSE progressions across the 45-hour forecast horizon for wind (left) and solar (right) power prediction.}
\label{fig:model_performance}
\end{figure}

The CNN-Transformer architecture demonstrated robust forecasting capabilities across the 45-hour prediction horizon. Figure \ref{fig:model_performance} shows the temporal evolution of forecasting errors for both wind and solar predictions. Solar forecasting exhibits pronounced diurnal error patterns that directly correlate with generation cycles with errors peak during midday hours. Wind forecasting shows less pronounced temporal patterns, reflecting the more stochastic nature of wind resources. Nevertheless, both models maintain relatively stable performance across the forecast horizon with limited systematic error degradation over extended lead times.

\subsection{Comparison with Operational Baseline}

The obtained predictions were benchmarked against ENTSO-E operational forecasts to evaluate their performance. Since this work involved a two-day forecast, the first-day predictions were compared with ENTSO-E's intraday predictions, and the second-day predictions were compared with ENTSO-E's day-ahead predictions. It is important to note that the CNN-Transformer forecasts use weather inputs initialized once before 3 a.m. on the forecast day, while ENTSO-E intraday predictions are provided at least once before 8 a.m., and day-ahead predictions at 6 p.m. Thus, the CNN-transformer model operates under an initialization disadvantage.

Table \ref{tab:performance} presents comprehensive performance metrics. For wind forecasting, there was an RMSE reduction from 205.37 MW to 76.06 MW, as well as an R² improvement from -17.02 to 0.19. The negative R² value of ENTSO-E wind forecasts indicates that their performance is worse than that of climatological averages. For solar forecasting, the RMSE decreased from 92.89 MW to 73.78 MW, and the R² increased from 0.78 to 0.89. The consistent 20\% improvement across metrics demonstrates a systematic enhancement.

\begin{figure}[tb]
\centering
\begin{minipage}{0.48\textwidth}
\centering
\includegraphics[width=\textwidth]{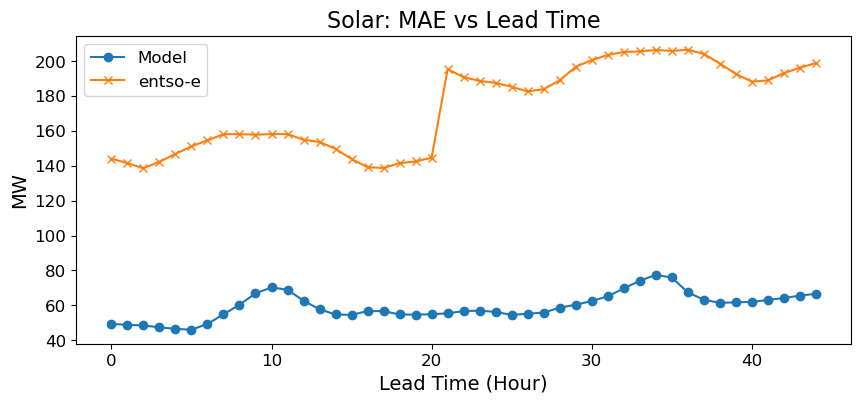}
\end{minipage}
\hfill
\begin{minipage}{0.48\textwidth}
\centering
\includegraphics[width=\textwidth]{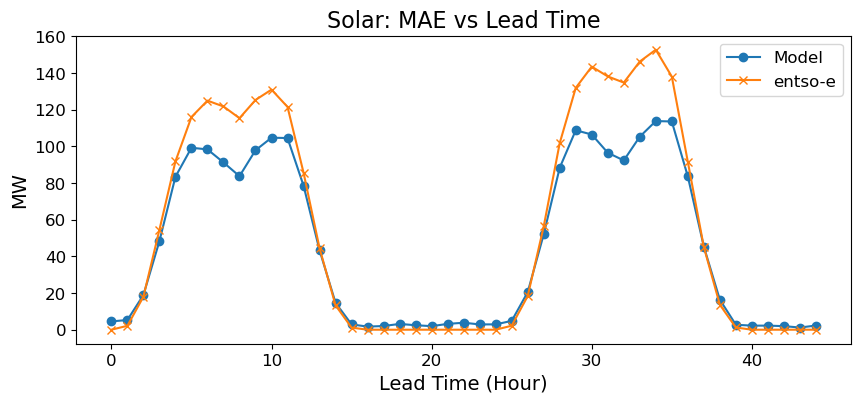}
\end{minipage}
\caption{Hourly MAE for wind (left) and solar (right) over the 45-hour horizon with the proposed model (blue) and ENTSO-E (orange)}
\label{fig:mae_comparison}
\end{figure}

Figure \ref{fig:mae_comparison} shows that the CNN-Transformer model demonstrates consistent hourly MAE patterns across energy sources. In contrast, ENTSO-E forecasts demonstrate highly variable and consistently greater bias magnitudes, particularly for wind, where the increase from day one to day two is due to different initialization of the forecast. This superior bias stability indicates that the CNN-Transformer model has more reliable operational deployment characteristics.

\begin{table}[tb]
\centering
\caption{Performance comparison between proposed model and ENTSO-E operational forecasts.}
\label{tab:performance}
\resizebox{\textwidth}{!}{%
\begin{tabular}{lcccccc}
\hline
\textbf{Metric} & \textbf{Wind Model} & \textbf{Wind ENTSO-E} & \textbf{Wind Improve} & \textbf{Solar Model} & \textbf{Solar ENTSO-E} & \textbf{Solar Improve} \\ \hline
RMSE {[}MW{]} & 76.06 & 205.37 & 62.96\% & 73.78 & 92.89 & 20.57\% \\
MAE {[}MW{]} & 59.43 & 173.54 & 65.75\% & 45.85 & 55.15 & 16.86\% \\
R² & 0.19 & -17.02 & - & 0.89 & 0.78 & 14.10\% \\
CVRMSE {[}\%{]} & 36.01 & 97.35 & 63.01\% & 35.62 & 44.85 & 20.58\% \\
NRMSE {[}\%{]} & 7.86 & 21.24 & 62.99\% & 6.15 & 7.75 & 20.65\% \\ \hline
\multicolumn{7}{c}{Target data range: Wind 0-967 MW (mean 211 MW, std 216 MW); Solar 0-1198 MW (mean 207 MW, std 270 MW)}
\end{tabular}%
}
\end{table}

\section{Discussion}

The demonstrated forecasting improvements translate directly to climate impact through two critical pathways. First, better renewable predictions enable grid operators to reduce fossil fuel backup generation by decreasing uncertainty margins that necessitate carbon-intensive reserve capacity. Second, improved forecasting reduces renewable energy curtailment, which is the practice of disconnecting renewable sources when generation exceeds predicted demand.


Importantly, the "weather maps as tokens" paradigm accomplishes these improvements in a computationally efficient manner with the same architecture for different sources. Training requires only one to two hours on a single NVIDIA A100 GPU and has 274,000 trainable parameters and a 1.1 MB model size. The modular design enables rapid regional adaptation. Spatial encoding can accommodate different geographical areas, and the temporal model remains unchanged, which facilitates scalability.

However, limitations include validation only for Sardinia, necessitating regional adaptation with local data, and forecast quality being inherently limited by the accuracy of the underlying NWP. Additionally, the current focus on aggregate regional production may necessitate modifications for individual plant-level applications.

\section{Conclusion}

This work introduced a novel approach to renewable energy forecasting, treating weather maps as tokens in transformers and bridging the paradigms of computer vision and natural language processing. The proposed CNN-Transformer architecture substantially outperformed operational baselines, reducing wind error by 63\% and improving solar forecasting by 21\%.

These improvements directly translate to climate impact by enhancing the integration of renewable energy into the electrical grid. Regardless of the specific renewable energy source, the lightweight architecture and modular design facilitate widespread deployment, which is particularly valuable for scaling renewable forecasting capabilities.


\section{Acknowledgements}
This work was developed in the Weather 4 Energy \& Infrastracture project funded by Italian Research Center on High Performance Computing Big Data and Quantum Computing (ICSC), project funded by European Union - NextGenerationEU - and National Recovery and Resilience Plan (NRRP) - Mission 4 Component 2.

The author would like to thank Illumia S.p.A. for their active role in developing this approach, drawing on their extensive expertise in meteorological modeling and the energy sector.

The author would also like to thank Giacomo Masato, Ivan Gentile, Matteo Angelinelli, Noemi Ambrosi, and Matteo Lippi for their valuable feedback and discussions that helped shape this work.


\section*{Appendix}

Individual day forecasting examples were selected systematically (15th of every other month) to illustrate model performance across diverse seasonal conditions in 2024. This systematic sampling approach avoids selection bias while providing seasonal diversity across winter, spring, summer, and fall conditions. These examples demonstrate the model's robustness across varied meteorological conditions and seasonal patterns.

\begin{figure}[htb]
\centering
\includegraphics[width=\textwidth]{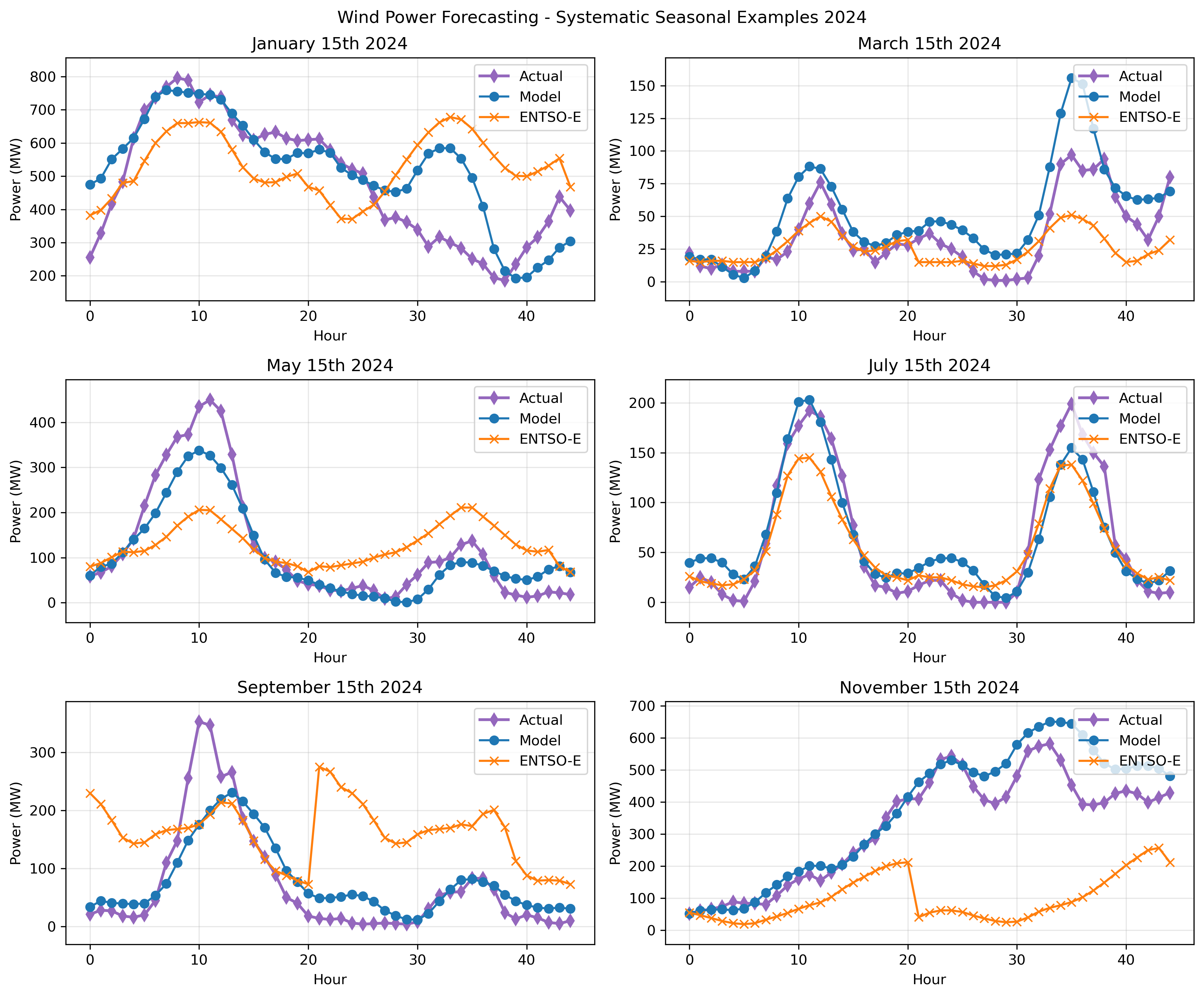}
\caption{Wind power forecasting across systematic seasonal examples in 2024. The CNN-Transformer model (blue circles) consistently demonstrates superior tracking of actual wind generation (purple diamonds) compared to ENTSO-E operational forecasts (orange crosses) across diverse seasonal and meteorological conditions.}
\label{fig:wind_examples}
\end{figure}

\begin{figure}[htb]
\centering
\includegraphics[width=\textwidth]{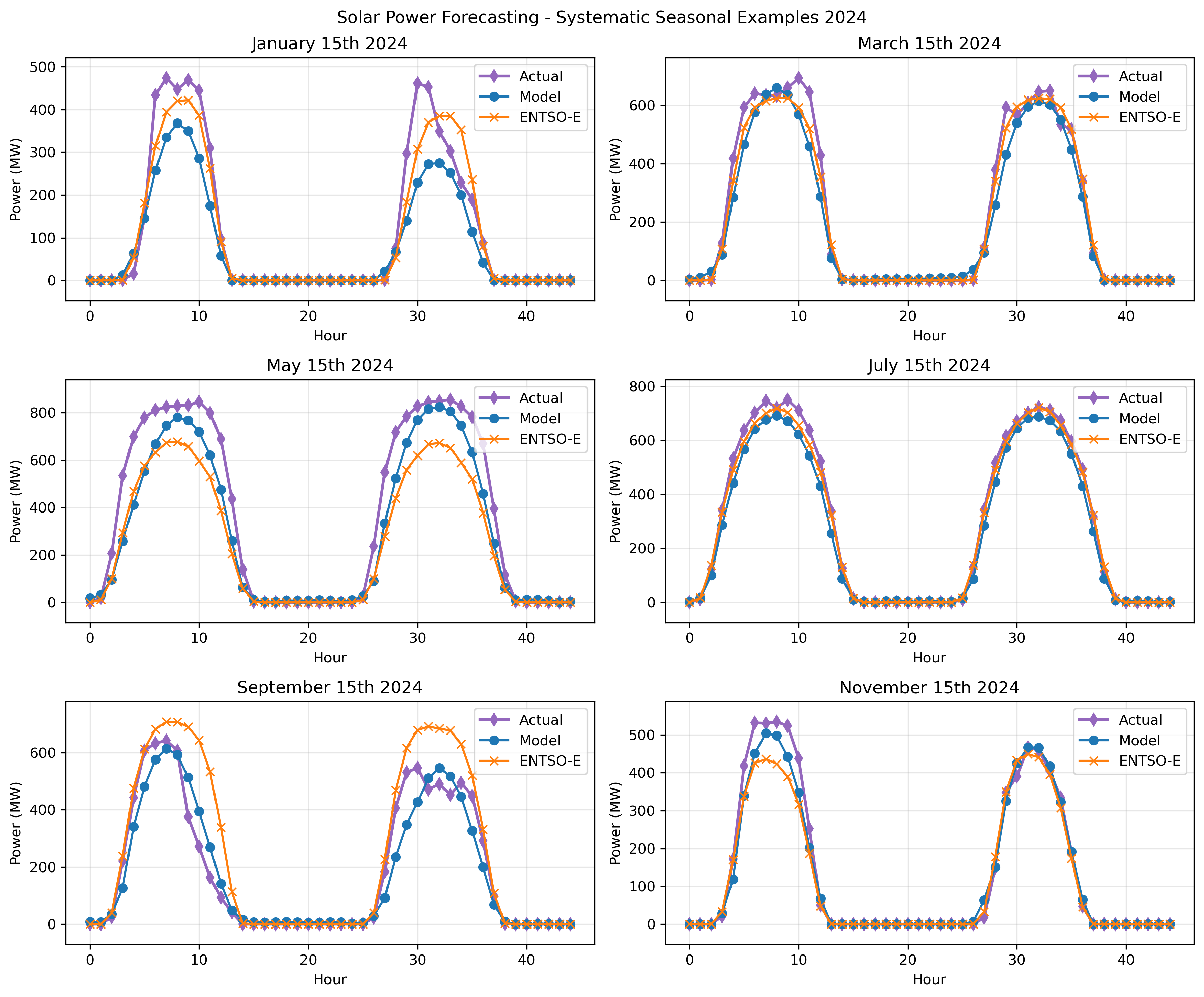}
\caption{Solar power forecasting across systematic seasonal examples in 2024. The model accurately captures diurnal solar patterns across all seasons, from reduced winter generation (January, November) to peak summer output (May, July). The CNN-Transformer model maintains closer alignment with actual generation compared to ENTSO-E forecasts. September 15th illustrates successful modeling of variable cloud conditions affecting solar irradiance.}
\label{fig:solar_examples}
\end{figure}

\end{document}